\documentclass{article}

\usepackage{arxiv}
\usepackage{subcaption}
\usepackage{booktabs} 
\usepackage{multirow}
\usepackage{algorithm}  
\usepackage{algorithmic} 

\usepackage[utf8]{inputenc} 
\usepackage[T1]{fontenc}    
\usepackage{hyperref}       
\usepackage{url}            
\usepackage{booktabs}       
\usepackage{amsfonts}       
\usepackage{nicefrac}
\usepackage{natbib}
\usepackage{microtype}      
\usepackage{lipsum}
\usepackage{graphicx}
\usepackage[dvipsnames]{xcolor}
\graphicspath{ {./images/} }

\usepackage{amsmath}
\usepackage{amssymb}
\usepackage{mathtools}
\usepackage{amsthm}
\usepackage[dvipsnames]{xcolor}
\usepackage[capitalize,noabbrev]{cleveref}

\theoremstyle{plain}

\theoremstyle{definition}

\theoremstyle{remark}

\usepackage[textsize=tiny]{todonotes}

\title{Epistemic Wrapping for Uncertainty Quantification}

\author{
  Maryam Sultana \\
  Oxford Brookes University, UK \\
  \texttt{msultana@brookes.ac.uk} \\
  \And
  Neil Yorke-Smith \\
  Delft University of Technology, The Netherlands \\
  \texttt{n.yorke-smith@tudelft.nl } \\
  \And
  Kaizheng Wang \\
  KU Leuven University, Belgium \\
  \texttt{kaizheng.wang@kuleuven.be } \\
  \And
  Shireen Kudukkil Manchingal \\
  Oxford Brookes University \\
  \texttt{skudukkil-manchingal@brookes.ac.uk} \\
  \And
  Muhammad Mubashar \\
  Oxford Brookes University \\
  \texttt{mmubashar@brookes.ac.uk} \\
  \And
  Fabio Cuzzolin \\
  Oxford Brookes University \\
  \texttt{fabio.cuzzolin@brookes.ac.uk} \\
}

\begin{document}
\maketitle
\begin{abstract}
Uncertainty estimation is pivotal in machine learning, especially for classification tasks, as it improves the robustness and reliability of models. We introduce a novel `Epistemic Wrapping' methodology aimed at improving uncertainty estimation in classification. Our approach uses Bayesian Neural Networks (BNNs) as a baseline and transforms their outputs into belief function posteriors, effectively capturing epistemic uncertainty and offering an efficient and general methodology for uncertainty quantification. Comprehensive experiments employing a Bayesian Neural Network (BNN) baseline and an Interval Neural Network for inference on the MNIST, Fashion-MNIST, CIFAR-10 and CIFAR-100 datasets demonstrate that our Epistemic Wrapper significantly enhances generalisation and uncertainty quantification.
\end{abstract}

\section{Introduction}
\label{sec:intro}
\vspace{-2pt}

In the realm of machine learning, particularly in classification tasks, uncertainty estimation plays a crucial role in enhancing the robustness and reliability of models \citep{sale2023volume}. Accurately quantifying uncertainty is vital for applications where decisions must be made with confidence, such as in medical diagnosis \citep{lambrou2010reliable}, autonomous driving \citep{fort2019large} and financial forecasting. Traditional deterministic neural networks, while powerful, cannot often effectively capture and express uncertainty \citep{liu2020simple}. 
This shortfall has spurred interest in probabilistic approaches, with Bayesian neural networks (BNNs) emerging as a promising solution in this context.  BNNs offer a principled approach to uncertainty estimation by incorporating prior distributions over the model parameters, leading to posterior distributions that reflect model uncertainty \citep{jospin2022hands}.  Despite their theoretical appeal, BNNs face practical challenges, including high computational costs and complexity in training.  

\vspace{-2pt}
The literature majors on two sources of uncertainty: \textit{Epistemic} Uncertainty (EU) and \textit{Aleatoric} Uncertainty (AU) \citep{hullermeier2021aleatoric, abdar2021review}. Epistemic uncertainty is due to a lack of knowledge about the true model parameters and can be reduced with more data or better models. In contrast, aleatoric uncertainty (AU) stems from the inherent randomness in the data generation process and cannot be reduced.
Over the years, various studies \citep{hullermeier2021aleatoric,abdar2021review} 
have recognised
that accurately modelling parameter uncertainty can produce a variety of credible network models, 
which are likely to include the true underlying network model,
leading to both better EU estimation and more reliable inference.
In particular, \emph{second-order uncertainty} frameworks 
(including belief functions \cite{cuzzolin2020geometry})
can be employed to model both EU and AU, effectively expressing `uncertainty about a prediction's uncertainty' \citep{hullermeier2021aleatoric,sale2023volume}.

\begin{figure}[t!]
\centering
\includegraphics[width=0.5\textwidth]{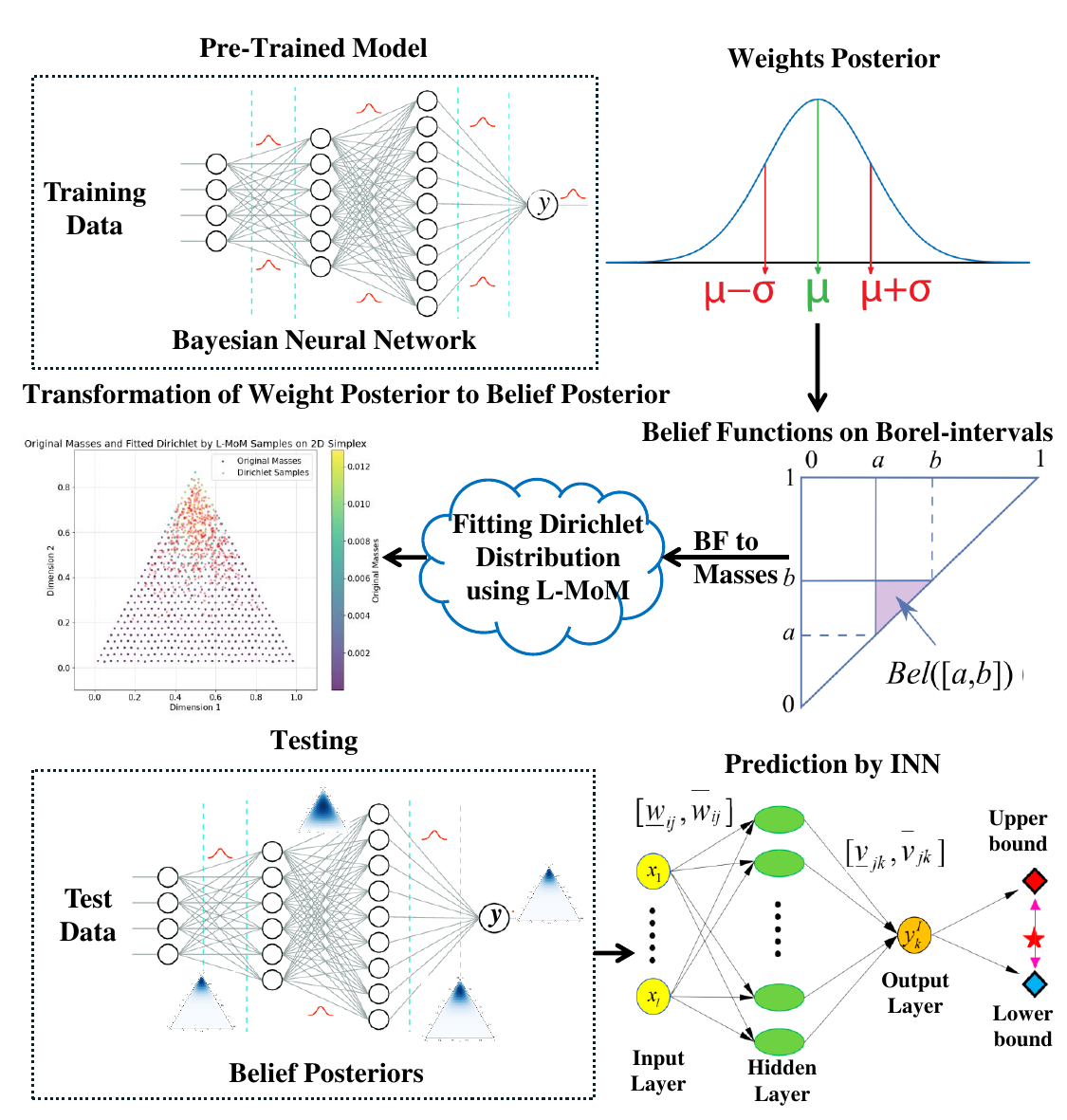}
\caption{Epistemic Wrapper transforms weights posteriors from a Bayesian Neural Network into belief posteriors through a five-step process. It involves extracting probability posteriors, calculating belief values over Borel intervals, computing mass values using Moebius inversion, fitting a Dirichlet distribution to these masses via method of L-moments, and using the resulting belief posteriors as weights to Interval Neural Networks (INNs) for final predictions. }

\label{fig: proposed_method}
\end{figure}

BNNs, as one of the prevalent method for uncertainty estimation, treat all the weights and biases of the network as probability distributions.  The prediction of the NN is represented as a second-order distribution, thus representing the probability distribution of distributions \citep{hullermeier2021aleatoric}. 
Although effective approximation techniques have been developed, such as variational inference approaches \citep{blundell2015weight,gal2016dropout} and sampling methods \citep{neal2011mcmc,hoffman2014no}, the high computational cost of BNNs during training as well as inference time limit their practical adoption, especially in real-time applications \citep{abdar2021review}.

\vspace{-2pt}
Recent results support the claim that modelling EU using uncertainty measures more general than probability distributions \cite{cuzzolin2024uncertainty}, such as credal sets \cite{levi1980enterprise} or random sets / belief functions \cite{shafer1976mathematical}, can lead to better uncertainty estimation and robustness \cite{manchingal2022epistemic, RS-NN, manchingal2023random, chan2024estimating, wang2024credal,wang2024credalwrapper,caprio2024credal,manchingal2025unifiedevaluationframeworkepistemic}.
 Still, all those efforts model (epistemic) uncertainty in the model's \emph{target} space, rather than its \emph{parameter} space. 
To our knowledge, no attempts have been made to model EU in the parameter space via higher-order uncertainty measures.

\vspace{-2pt}
This paper proposes
\emph{Epistemic Wrapper}, a novel method 
which, for the first time, models EU in the parameter space via a random set representation by ``wrapping" a learnt Bayesian posterior there in the form of 
a \emph{belief function}.
posterior (Fig.~\ref{fig: proposed_method}). 
Our {Epistemic Wrapper}
follows a structured five-step process,  
where each step is executed in a hierarchical manner: (i) We begin by extracting posterior distributions with parameters ($\mu, \sigma$) from a pre-trained BNN model, where the priors are modeled as Gaussian distributions. 
(ii) The posterior distributions are then truncated, and continuous belief functions are computed over closed intervals. (iii) In the third step, these belief values are transformed into mass values using Moebius inversion.  
(iv) A Dirichlet distribution is fitted to the grid of mass values using the Method of L-moments.  
(v) Finally, inference is performed using a Hybrid Interval Neural Network (INN).
\vspace{-2pt}
Our approach leverages the strengths of BNNs while injecting the ability of higher-order measure to improve robustness and uncertainty estimation. 

\vspace{-2pt}
Our contributions are therefore:
\begin{enumerate}
\vspace{-5pt}
\item \textbf{A first attempt to model EU in the parameter space using higher-order uncertainty measures}.
\item This happens via a new, versatile \textbf{Epistemic Wrapper} concept,
that can be applied to any BNN baseline to convert it automatically into a belief-function posterior. 
\item Based on the above, a \textbf{novel approach to uncertainty estimation}
in classification which efficiently leverages BNNs as a foundation. 
\end{enumerate}


\vspace{-5pt}
Our experiments demonstrate the versatility of the proposed Epistemic-wrapper approach across the BNN baseline on two datasets: MNIST and Fashion-MNIST. The results indicate that the epistemic wrapper generalizes effectively across these diverse datasets, significantly improving performance over the baseline BNN. On MNIST dataset, the baseline BNN achieved an accuracy of 72.44\% $\pm$ 0.24, whereas the Epi-Wrapper substantially outperformed it, achieving 91.02\% $\pm$ 0.05. Similarly, on the Fashion-MNIST dataset, the baseline BNN attained an accuracy of 58.91\% $\pm$ 0.24, while the Epi-Wrapper demonstrated a notable improvement, reaching 82.45\% $\pm$ 0.10. These results highlight the effectiveness of our approach in enhancing predictive performance across different datasets.

\textbf{Why do we model epistemic uncertainty in the parameter space rather than in the target space?}\\
The motivation behind modeling epistemic uncertainty in the parameter space using higher-order uncertainty measures stems from the idea that parameter uncertainty is a primary source of epistemic uncertainty. By modeling uncertainty directly in the parameter space before it propagates to predictions we capture model-level uncertainty in a more principled way. Modeling in the parameter space offers several advantages: (a) It provides a prior-agnostic mechanism to represent epistemic uncertainty, without relying solely on the model’s output distribution. (b) It enables structured and interpretable sampling through belief functions and Dirichlet distributions, supporting more stable and calibrated uncertainty estimates via interval-based inference. (c) It can be seamlessly integrated with existing BNNs without retraining, offering flexibility and broader applicability.
\\
The paper is organised as follows.  Section~\ref{related_work} surveys the relevant literature. 
Section~\ref{Epistemic_Wrapper} explains in detail our epistemic wrapper approach. 
Section~\ref{Experiments} reports the experiments and results.  Section~\ref{conclusion} concludes.


\vspace{-5pt}
\section{Relevant Work}
\label{related_work}\label{sec:related}
\vspace{-2pt}


\subsection{Bayesian Neural Networks}
Bayesian Neural Networks (BNNs) provide a principled framework for uncertainty estimation by treating weights and biases as probability distributions, supporting robust decision-making \citep{jospin2022hands}. Efficient methods like Variational Inference (e.g., Bayes by Backprop \citep{blundell2015weight}) and dropout-based approximations \citep{gal2016dropout} have made BNNs scalable. However, their high computational demands, especially during training and inference, remain a significant challenge for practical applications \citep{abdar2021review}.


\vspace{-2pt}
While various types of uncertainty measures \cite{cuzzolin2021big} have been employed in machine learning in the past \cite{cuzzolin2013if,cuzzolin18belief-maxent,liu2019evidence,gong2017belief},
recent advancements in epistemic uncertainty modelling have introduced a range of methods to improve predictive reliability across various neural architectures. Evidential deep learning predicts second-order probability distributions to estimate uncertainty, but faces challenges in optimisation and interpretation \citep{juergens2024is}. Methods like G-$\Delta$UQ refine uncertainty calibration in Graph Neural Networks (GNNs) through stochastic data centering \citep{trivedi2024accurate}, while SPDE-based GNNs employ $Q$-Wiener processes for uncertainty propagation in complex graphs \citep{lin2024graph}. The Graph Energy-Based Model (GEBM) leverages graph diffusion to quantify uncertainty at different structural levels \citep{fuchsgruber2024energybased}, and credal set-based ensemble learning constructs plausible probability distributions to measure aleatoric and epistemic uncertainty \citep{hofman2024quantifying}.

Crucially, 
\citep{manchingal2025unifiedevaluationframeworkepistemic} introduces a unified evaluation framework for uncertainty-aware classifiers, mapping all uncertainty-aware predictions into credal sets \cite{cuzzolin2008credal}, thus enabling a standardised assessment of epistemic uncertainty across BNNs, Deep Ensembles, Evidential Deep Learning (EDL), and Credal Set-based approaches. 
\citep{RS-NN} extends uncertainty modelling through Random-Set Neural Networks (RS-NNs), which employ random set theory to construct belief-based uncertainty representations, providing a more flexible alternative to conventional probabilistic models. 
Credal Interval Neural Networks 
\cite{wang2025creinns}, instead, 
represent predictions as credal sets, which encapsulate a range of probable outcomes, thereby explicitly modelling epistemic uncertainty. Building on the latter,
Credal Deep Ensembles 
\citep{wang2024credal} 
predict and aggregate ensembles of convex sets of probability distributions, 
resulting in a more conservative and informative epistemic uncertainty quantification. 
In an alternative approach
\cite{charpentier2020posterior, malinin2019ensemble, sensoy2018evidential} predictions are modelled as Dirichlet distributions. A key challenge with these methods is the lack of ground truth labels for uncertainty, making direct supervision difficult.

While these models can be highly effective, they primarily quantify uncertainty at the target level, leaving 
the question of modelling epistemic uncertainty at parameter level open.
In contrast, our proposed 
Epistemic Wrapper
leverages BNNs to do exactly so, 
by transforming probability posteriors into belief posteriors, 
to offer a robust solution for uncertainty quantification in classification tasks. 

\vspace{-2pt}
\subsection{Interval neural networks}

\vspace{-2pt}
Traditional \textit{Interval Neural Networks} (INNs) employ deterministic interval-based representations for inputs, outputs, weights, and biases, ensuring robust uncertainty modelling in neural computations. The forward propagation in an INN follows interval arithmetic principles, where the interval-formed activations in each layer are computed using element-wise interval addition, subtraction, and multiplication \citep{hickey2001interval}. Specifically, the activation output of the $l^{\text{th}}$ layer is determined by applying a monotonically increasing activation function to the interval-weighted sum of the previous layer's outputs and the corresponding interval biases. This formulation guarantees the \emph{set constraint} property, ensuring that for any given input and network parameters within their defined intervals, the computed activations remain bounded within a well-defined range. When the activation function is non-negative (e.g., ReLU), further simplifications allow efficient computation of interval bounds using minimum and maximum operators. This structured interval propagation enables INNs to maintain rigorous mathematical constraints while modelling uncertainties in deep learning architectures \citep{ASPINN}. 

\vspace{-2pt}
In our approach, we employ INNs at inference time using our epistemic wrapper weights, sampled from the wrapped belief posterior, thus leveraging their structured interval-based representations to quantify and propagate epistemic uncertainty effectively. 

\vspace{-5pt}
\section{Methodology}
\label{Epistemic_Wrapper}\label{sec:method}
\vspace{-2pt}

The proposed Epistemic Wrapper approach consists of a five-step process for transforming learnt posterior distributions on model parameters into belief functions posteriors.

\vspace{-2pt}
\subsection{Learning a Bayesian Posterior (Baseline)}

\vspace{-2pt}
In the first step we use BNNs trained on Gaussian priors with parameters $(\mu, \sigma)$ and obtain posterior distributions with the same parameters. The posterior distribution $p(\boldsymbol{\omega}|\mathbb{D})$ is learned using Bayes' theorem, where `Variational Inference’ (VI) is employed to approximate the intractable posterior by optimizing a variational distribution $q(\boldsymbol{\omega})$ that closely matches $p(\boldsymbol{\omega}|\mathbb{D})$. During inference, we approximate the posterior via Bayesian Model Averaging by sampling weights from the variational posterior.

\vspace{-2pt}
\subsection{Dynamic Truncation} 

\begin{figure}[ht!]
\centering
\includegraphics[width = 0.5\textwidth]{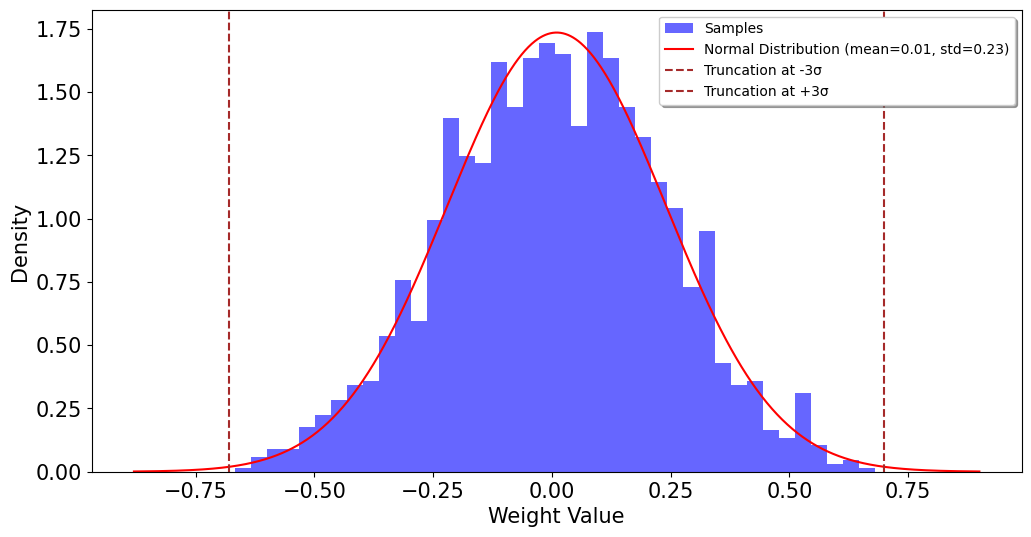}

\caption{Posterior distribution of weights 
for the last layer. The histogram displays the sampled weights, overlaid with a fitted normal distribution (mean = 0.01, std = 0.23). Vertical dashed lines indicate truncation points at $\pm 3\sigma$.} \label{weights_distribution}

\end{figure}

\vspace{-2pt}
A typical Bayesian weight posterior will look like the one in Figure~\ref{weights_distribution}. 

In the second step, 
such posterior distributions are truncated using a \emph{dynamic distribution truncation} mechanism, designed as an adaptive technique used to define the range of a distribution based on its mean and standard deviation. This dynamically scales the bounds to parameter values according to the variance of the distribution, 
ensuring
that the truncation bounds are tighter for distributions with smaller variances, as they inherently have more concentrated probability masses, and looser for those with larger variances. 

\vspace{-2pt}
The truncation bounds are calculated as:
$ \text{Lower Bound} = \mu - \text{dynamic\_multiplier} \cdot \sigma, \quad \text{Upper Bound} = \mu + \text{dynamic\_multiplier} \cdot \sigma,
$
where \(\mu\) is the mean, \(\sigma\) is the standard deviation, and the \(\text{dynamic\_multiplier}\) is calculated as:
$  \text{dynamic\_multiplier} = \min(5.0, \frac{1.0}{\sigma}),$ ensuring that the multiplier decreases for low-variance distributions while capping its value at 5.0 to prevent excessive truncation in high-variance cases. We have selected this approach as it provides a balance between capturing the significant probability mass of the distribution and avoiding overly wide or narrow bounds, which could either dilute meaningful mass representation or exclude critical probabilistic regions. 

\vspace{-2pt}
\subsection{Continuous Belief Functions on Closed Intervals} 

\vspace{-2pt}
\textbf{Belief Functions}. Belief functions, grounded in the mathematical framework of random sets, were initially introduced by Dempster \citep{dempster2008upper} and later formalized by Shafer \citep{Shafer76} as an alternative model for subjective belief to Bayesian probability. 
In finite domains, such as a collection of classes, belief functions are characterised by a \emph{basic probability assignment} (BPA) \citep{Shafer76}, which is a set function $m: 2^\Theta \to [0,1]$ satisfying $m(\emptyset) = 0$ and $\sum_{A \subseteq \Theta} m(A) = 1$. The value $m(A)$ is interpreted as the probability mass directly assigned to subset $A \subseteq \Theta$ in a random-set formulation \citep{smets91other}.  
Subsets $A$ of $\Theta$ with $m(A) > 0$ are referred to as \emph{focal elements}. 
Classical belief functions extend the notion of discrete mass functions by assigning normalized, non-negative mass values not only to elements $\theta \in \Theta$ but to subsets of $\Theta$, governed by:
\begin{equation}
m(A) \geq 0, \forall A \subseteq \Theta, \quad \sum_{A \subseteq \Theta} m(A) = 1.
\end{equation}
The belief function $Bel(A)$ associated with a mass function $m$ is defined as the total mass assigned to all subsets $B \subseteq A$. Conversely, $m$ can be recovered from $Bel$ through Moebius inversion~\cite{Shafer76}:
\begin{equation}
Bel(A) = \sum_{B \subseteq A} m(B), \quad m(A) = \sum_{B \subseteq A} (-1)^{|A \setminus B|} Bel(B).
\end{equation}
This formulation demonstrates that classical probability measures are a special case of belief functions, assigning mass exclusively to singletons.

\vspace{-2pt}
\textbf{Continuous belief functions on intervals}. Belief functions can be easily extended to continuous spaces (e.g., a network's parameter space) by defining a continuous mass function over the collection of \emph{closed intervals}, rather than the entire power set.
Given a network parameter $\omega$ with values in $\mathbb{R}$, this requires defining a continuous PDF over the collection of intervals $[a,b] \subset \mathbb{R}$ \citep{cuzzolin2020geometry}.
Here we will assume that parameter values are bounded after truncation (for illustration, in $[0,1]$); however, the method can be easily extended to unbounded parameter values as well.

\begin{figure}[tb]
\centering
\includegraphics[width =0.5\textwidth]{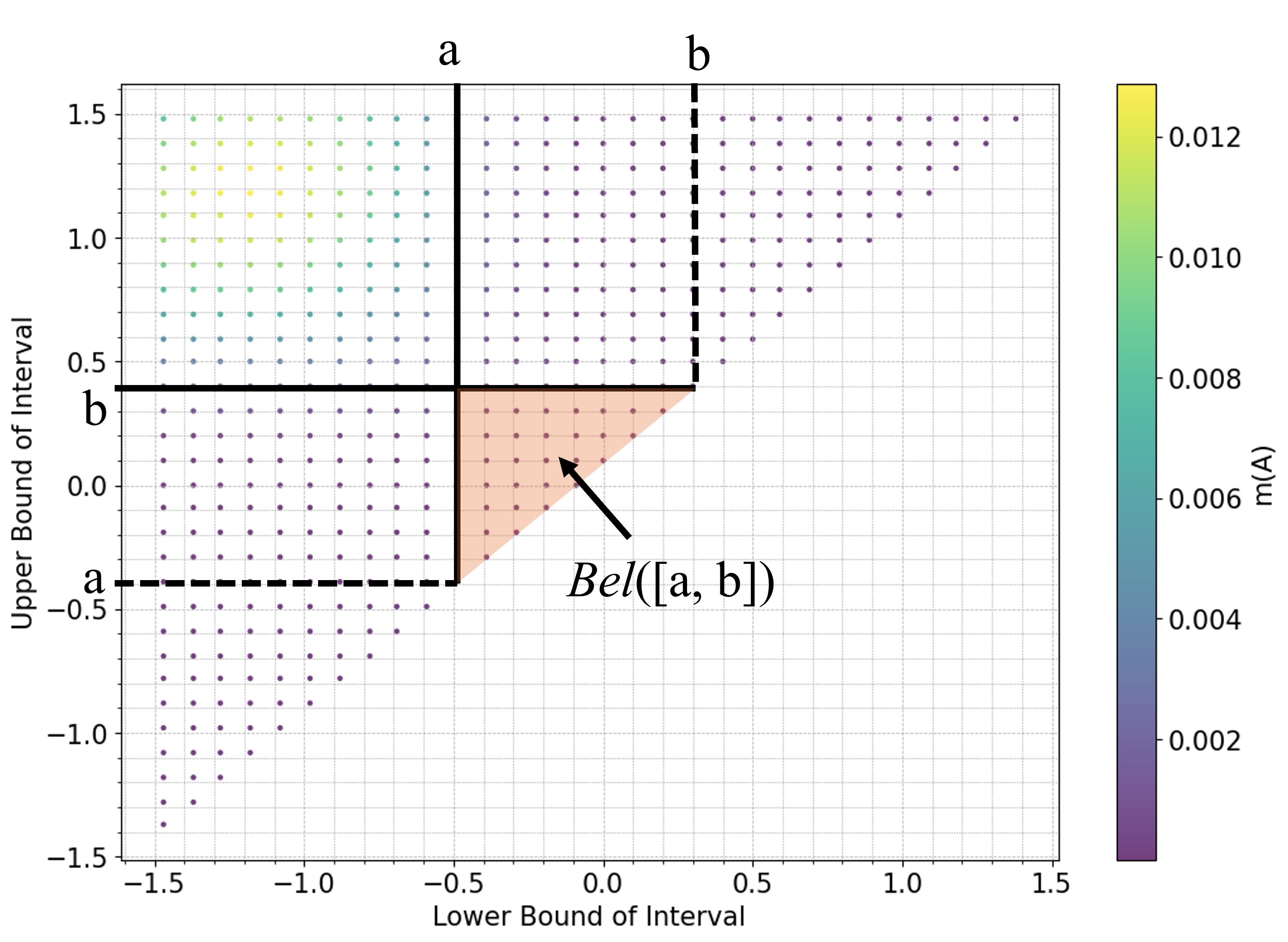}

\caption{Graphical visualisation of the continuous PDF/mass function on intervals, with the area whose integral amounts to $Bel([a,b])$. 
} \label{borel_intervals}
\vspace{-12pt}
\end{figure}

\vspace{-2pt}
The space of all closed intervals in $[0,1]$ is a triangle, as illustrated in Fig.~\ref{borel_intervals}. Given a continuous mass function there (non-negative and with integral 1), 
one can compute the belief and plausibility value of a parameter interval $A = [a,b]$ by integrating it over specific regions of the triangle \cite{smets05real} (Fig. \ref{borel_intervals}). The same applies for parameters bounded by arbitrary values.

Given a posterior distribution over a network's weight, learned by a BNN, our Wrapper transforms it into a continuous belief function using the 
method proposed in \cite{Wasserman90}. 
For any closed interval $A = [a,b]$ of the parameter space, one can compute its \emph{plausibility} from the posterior distribution by taking the supremum of the normalised posterior $\hat{p}(\boldsymbol{\omega}|\mathbb{D})$ across all $\boldsymbol{\omega} \in A$, namely:
\begin{equation}
\vspace{-6pt}
    Pl_{\Theta}(A|\mathbb{D}) = \sup_{\omega \in A} \hat{p}(\boldsymbol{\omega}|\mathbb{D}).
    \vspace{-2pt}
\end{equation}
The corresponding belief value is then calculated as the complement of the plausibility:
\begin{equation}
\vspace{-6pt}
    Bel_{\Theta}(A|\mathbb{D}) = 1 - Pl_{\Theta}(A^{c}|\mathbb{D}),
    \vspace{-2pt}
\end{equation}
ultimately providing the sought random-set representation in the parameter space. 

\vspace{-2pt}
The method is grounded into rationality principles, such as (i) the likelihood principle, (ii) compatibility with Bayesian inference (which
ensures that combining a Bayesian prior with the belief function yields the Bayesian
posterior), and (iii) the principle of Minimum Commitment, which maintains that among
the belief functions satisfying the previous two principles, the one chosen should commit
to the least amount of information necessary \cite{cuzzolin2020geometry}.

\vspace{-2pt}
To cap complexity,
sample belief values can be computed for a grid of parameter values only. 
The corresponding mass values can then be easily obtained by Moebius inversion \citep{shafer1976mathematical}. 

\vspace{-2pt}
\subsection{Fitting a Dirichlet Distribution}
\vspace{-2pt}

The fourth step of our Epistemic wrapper employs the method of \emph{L-moments} \cite{L-Moments} to fit a Dirichlet distribution to the grid of mass values so obtained.

\begin{figure}[ht!]
\centering
\includegraphics[width = 0.45\textwidth]{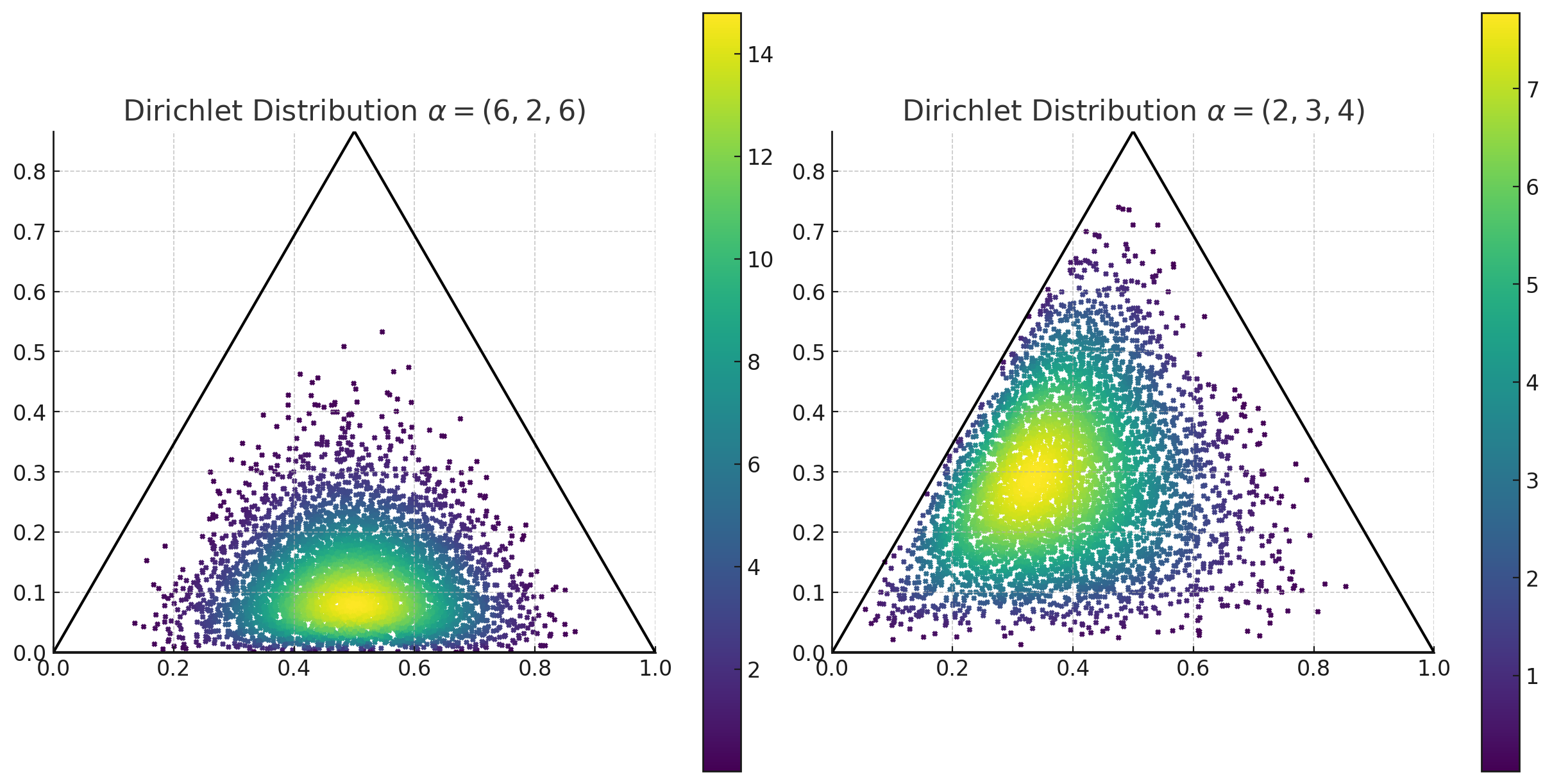}
\vspace{-8pt}
\caption{Probability densities of the Dirichlet distribution  as functions on the 2D-simplex: $\alpha$ = (6,2,6) (left), $\alpha$ = (2,3,4) (right). 
} \label{Dirichlet_distribution}
\vspace{-4pt}
\end{figure}

\vspace{-2pt}
A \textbf{Dirichlet distribution} is a family of continuous multivariate probability distributions parameterised by a vector $\alpha$ of positive real numbers; in
fact, a multivariate extension of the Beta distribution (Figure~\ref{Dirichlet_distribution}):
\begin{equation}
\vspace{-8pt}
    f(x_1, \ldots, x_K; \alpha_1, \ldots, \alpha_K) = \frac{1}{B(\alpha)} \prod_{i=1}^K x_i^{\alpha_i - 1}.
    \vspace{-1pt}
\end{equation}
As they are defined on the collection of vectors $x \in [0,1]^K$ of dimension $K$ whose coordinates add to 1,  Dirichlet distributions can be interpreted as second-order distributions.

The
\textbf{method of L-Moments}
is a statistical approach employed for parameter estimation in probability distributions. Here we utilise this method to estimate the parameters of a Dirichlet distribution over mass values. 
L-moments are analogous to conventional moments but are based on linear combinations of order statistics, making them more robust to outliers and capable of providing a more reliable characterization of the data. 

\begin{figure}[ht!]
\centering
\includegraphics[width = \columnwidth]{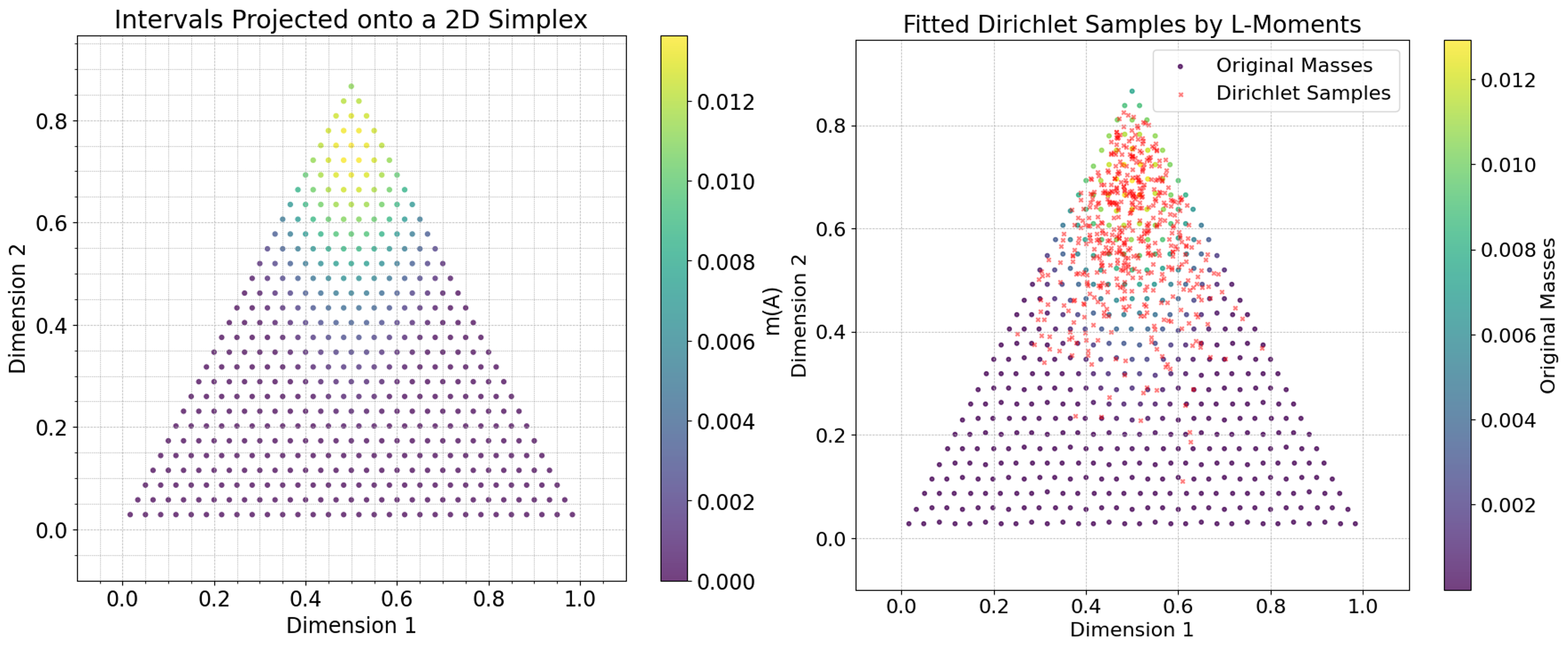}
\vspace{-8pt}
\caption{\textbf{Left:} Scatter plot showing intervals projected onto a 2D simplex. Each point represents an interval $A = [a,b]$ with its location determined by 
the values $a$ and $b$, and the colour scale indicates the corresponding mass values $m(A)$, ranging from 0.00 to 0.012. \textbf{Right:} Visualization of Dirichlet samples on a 2D simplex. The scatter plot shows points sampled from the fitted Dirichlet distribution over mass values. 
} \label{2D_to_3D_simplex}
\vspace{-5pt}
\end{figure}

\vspace{-2pt}
To fit a Dirichlet distribution to the grid of mass values, we compute weighted L-moments from the data represented in a 3D simplex space, 
where each data point has an associated weight derived from its mass value.
An example grid
in a 3D simplex representation is shown in Figure~\ref{2D_to_3D_simplex}-Left. 

\vspace{-2pt}
\textbf{Computation of weighted L-Moments}. 
We first need to compute
the first-order and second-order weighted L-moments from the grid of data points.
Let $\mathbf{x}_i \in \mathbb{R}^3$ denote the $i$-th data point in the 3D simplex and $w_i$ its associated weight (derived by normalizing the mass values, so that $\sum_i w_i = 1$). The L-moments are computed as follows.
\vspace{-2pt}
\emph{First-order L-moment ($L_1$)}. This is the weighted mean of the points in the simplex and is given by:
    \vspace{-4pt}
    \begin{equation}
        L_1 = \sum_{i=1}^n w_i \mathbf{x}_i.
    \end{equation}
\emph{Second-order L-moment ($L_2$)}. This represents the weighted spread (variance) of the points relative to $L_1$:
    \begin{equation}
        L_2 = \frac{\sum_{i=1}^n w_i (\mathbf{x}_i - L_1)^2}{\sum_{i=1}^n w_i}
    \end{equation}
To ensure numerical stability, a small value $\epsilon$ is added to $L_2$ when necessary, preventing division by zero in subsequent computations. 

\emph{Fitting the Dirichlet Distribution}. Using the computed L-moments, we can estimate the parameters $\boldsymbol{\alpha} = (\alpha_1, \alpha_2, \alpha_3)$ of the Dirichlet distribution. The relationship between L-moments and the Dirichlet parameters is expressed as: 
\begin{equation}
    \alpha_k = L_{1,k} \left( \frac{L_{1,k} (1 - L_{1,k})}{L_{2,k}} - 1 \right), \quad k = 1, 2, 3
\end{equation}
where $L_{1,k}$ and $L_{2,k}$ are the respective components of the first and second L-moments along each axis of the simplex.
The whole procedure is summarised in Algorithm~\ref{alg:l_moments}.

\vspace{-2pt}
As a sanity check, after fitting a Dirichlet distribution to the grid of mass values, samples from it are also concentrated on the top of the simplex as shown in Figure~\ref{2D_to_3D_simplex}-Right.

\vspace{-2pt}
\begin{algorithm}[tb]
\caption{Weighted L-Moments for Dirichlet Parameter Estimation}
\label{alg:l_moments}
\begin{algorithmic}[1]
\REQUIRE $\mathbf{X} \in \mathbb{R}^{n \times 3}$ (3D simplex coordinates), $\mathbf{w} \in \mathbb{R}^n$ (masses), $\epsilon > 0$
\STATE Normalize weights: $\mathbf{w} \gets \mathbf{w} / \sum_{i=1}^n w_i$
\STATE Compute $L_1 \gets \sum_{i=1}^n w_i \mathbf{x}_i$
\STATE Compute $L_2 \gets \frac{\sum_{i=1}^n w_i (\mathbf{x}_i - L_1)^2}{\sum_{i=1}^n w_i}$
\STATE Adjust $L_2 \gets \max(L_2, \epsilon)$
\STATE Estimate $\boldsymbol{\alpha} \gets L_1 \odot \left( \frac{L_1 \odot (1 - L_1)}{L_2} - 1 \right)$
\STATE Enforce positivity: $\boldsymbol{\alpha} \gets \max(\boldsymbol{\alpha}, \epsilon)$
\ENSURE Dirichlet parameters $\boldsymbol{\alpha}$
\end{algorithmic}
\end{algorithm}

\textbf{Theoretical Properties of the Epistemic Wrapper}\\
The Epistemic Wrapper preserves an important theoretical property. Specifically, the original Bayesian posterior \( P \) lies within the credal set induced by the belief and plausibility functions after wrapping, satisfying
\[
Bel(A) \leq P(A) \leq Pl(A) \quad \text{for all measurable sets } A.
\]
This relation ensures that our transformation is conservative: it enriches the original posterior with second-order uncertainty without distorting the underlying predictive information. Consequently, the model maintains consistency with the Bayesian posterior while gaining robustness, which helps to explain the observed improvements in generalization and uncertainty estimation. The plausibility \( Pl(A) \) captures the maximum value of \( \hat{p}(\omega|\mathbb{D}) \) over \( A \), while belief \( Bel(A) \) captures the minimum guaranteed mass by considering the complement \( A^c \).  
Since \( P(A) \) is the integral of \( \hat{p}(\omega|\mathbb{D}) \) over \( A \), it must lie between the least conservative estimate (Bel) and the most generous estimate (Pl) over \( A \).  
This follows from the construction rules of likelihood-based belief functions and random set theory (see \citep{Shafer76, Wasserman90, cuzzolin2020geometry}).

\vspace{-2pt}
\subsection{Budgeting}

\vspace{-2pt}
A budgeting strategy is introduced (detailed in Section \ref{sec:ablation}) to selectively transform the posterior distributions of a \emph{subset} of parameters (weights and biases). Posteriors that are not selected, referred to as \emph{unwrapped} posteriors, retain their original learned parameters. We propose four distinct budgeting strategies: three are parameter-based, prioritizing posteriors with high $\mu$, high $\sigma$, or simultaneously high $\mu$ and $\sigma$, while the fourth employs a random selection strategy that remains unbiased with respect to these parameter values.

\vspace{-2pt}
\subsection{Inference via Hybrid Interval Neural Networks}
\label{Subsec: ExistingINNs}
\vspace{-2pt}
The fifth step in our approach is inference.
For this purpose, we employ a Hybrid Interval Neural Network \emph{Hybrid-INN}, where weight intervals are derived from a combination of Dirichlet-derived intervals (wrapped parameters) and Gaussian posteriors (unwrapped parameters). Although our architecture is based on a standard INN, the way we handle these intervals is by computing the mean of the upper and lower bounds, which introduces an important distinction. For this reason, we refer to our model as Hybrid-INN. This averaging mechanism of the interval bounds prevents extreme values from distorting predictions. It ensures a more stable, well-calibrated uncertainty representation, particularly for epistemic uncertainty, which can otherwise be highly sensitive to interval width. It also harmonizes wrapped and unwrapped weights by smoothly integrating both Dirichlet and Gaussian-based uncertainty representations. 

Namely, the unwrapped weights (Gaussian posteriors) generate the following intervals:
$\text{Lower bound} = \mu - \sigma, \text{Upper bound} = \mu + \sigma.$ The process of defining these wrapped and unwrapped weights can be considered the weight initialisation step for the Hybrid-INN model. In contrast, the baseline INN retains the default weight initialisation scheme \citep{kim1993understanding}. In other words, at inference time, the model operates as a Hybrid-INN but with two distinct sets of weights: (i) with random initialisation, (baseline INN) and (ii) with weights transformed by the Epi-Wrapper. Both models are also fine-tuned: the baseline with randomly initialized weights, while the Epi-Wrapper utilizes weights transformed through the wrapping strategy.
For a fair comparison, both the baseline and our Epi-Wrapper operate under identical functional settings.

\section{Experiments}
\label{Experiments}\label{sec:results}
\vspace{-2pt}

In this section, we present a comprehensive evaluation of our Epi-Wrapper approach through a series of experiments designed to assess its effectiveness in uncertainty estimation and predictive performance. We begin by describing the experimental setup, including datasets, model architecture, and ablation studies. We then compare our method against relevant baselines, followed by an analysis of key performance metrics.

\begin{table*}[!t]
\caption{Classification accuracies on MNIST under different budgeting criteria before and after fine-tuning for posterior weights. Results are from 15 runs. Best scores are presented in bold.
}
\label{combined_results}
\small
\resizebox{\linewidth}{!}{
\begin{sc}
\begin{tabular}{ccccccccc}
\toprule
\multirow{2}{*}{Budgeting} & \multirow{2}{*}{MLP Size} & \multicolumn{2}{c}{Before Fine-Tuning} & \multicolumn{3}{c}{After Fine-Tuning} \\
\cmidrule(lr){3-4} \cmidrule(lr){5-7}
 & & INN & Epi-Wrapper & BNN & INN & Epi-Wrapper  \\
\midrule
\multirow{3}{*}{$\uparrow \sigma$} 
& 2  & 9.11 $\pm$ 0.53 & \textbf{12.93 $\pm$ 0.75} & 33.14 $\pm$ 0.22 & 57.77 $\pm$ 0.81 & \textbf{62.43 $\pm$ 0.46} \\
& 4  & 10.44 $\pm$ 0.77 & \textbf{19.94 $\pm$ 0.47} & 40.19 $\pm$ 0.55 & 83.32 $\pm$ 0.24 & \textbf{85.17 $\pm$ 0.10} \\
& 8  & 9.33 $\pm$ 0.54 & \textbf{25.46 $\pm$ 1.57} & 72.44 $\pm$ 0.24 & \textbf{91.12 $\pm$ 0.08} & 91.08 $\pm$ 0.09 \\
\midrule
\multirow{3}{*}{$\uparrow \mu$} 
& 2  & 9.11 $\pm$ 0.53 & \textbf{10.63 $\pm$ 0.34} & 33.14 $\pm$ 0.22 & 57.77 $\pm$ 0.81 & \textbf{63.06 $\pm$ 0.47} \\
& 4  & 10.44 $\pm$ 0.77 & \textbf{18.13 $\pm$ 0.71} & 40.19 $\pm$ 0.55 & 83.32 $\pm$ 0.24 & \textbf{85.35 $\pm$ 0.06} \\
& 8  & 9.33 $\pm$ 0.54 & \textbf{51.33 $\pm$ 1.21} & 72.44 $\pm$ 0.24 & \textbf{91.12 $\pm$ 0.08} & 91.02 $\pm$ 0.05 \\
\midrule
\multirow{3}{*}{$\uparrow \mu + \sigma$} 
& 2  & 9.11 $\pm$ 0.53 & \textbf{10.45 $\pm$ 0.16} & 33.14 $\pm$ 0.22 & 57.77 $\pm$ 0.81 & \textbf{63.02 $\pm$ 0.55} \\
& 4  & 10.44 $\pm$ 0.77 & \textbf{18.55 $\pm$ 0.68} & 40.19 $\pm$ 0.55 & 83.32 $\pm$ 0.24 & \textbf{85.18 $\pm$ 0.07} \\
& 8  & 9.33 $\pm$ 0.54 & \textbf{51.31 $\pm$ 1.29} & 72.44 $\pm$ 0.24 & 91.12 $\pm$ 0.08 & \textbf{91.12 $\pm$ 0.07} \\
\midrule
\multirow{3}{*}{RS} 
& 2  & \textbf{9.11 $\pm$ 0.53} & 9.80 $\pm$ 0.00 & 33.14 $\pm$ 0.22 & 57.77 $\pm$ 0.81 & \textbf{64.84 $\pm$ 0.16} \\
& 4  & 10.44 $\pm$ 0.77 & \textbf{17.35 $\pm$ 0.27} & 40.19 $\pm$ 0.55 & 83.32 $\pm$ 0.24 & \textbf{85.45 $\pm$ 0.06} \\
& 8  & \textbf{9.33 $\pm$ 0.54} & 9.23 $\pm$ 0.64 & 72.44 $\pm$ 0.24 & \textbf{91.12 $\pm$ 0.08} & 90.80 $\pm$ 0.09 \\
\bottomrule
\end{tabular}
\end{sc}
}
\vskip -0.2in
\end{table*}

\vspace{-2pt}
\subsection{Implementation Details}
\vspace{-2pt}

\textbf{Datasets}. We evaluated the performance of the Epistemic Wrapper on three classification benchmarks: MNIST \citep{lecun1998mnist}, Fashion-MNIST \cite{xiao2017fashion} and CIFAR-10 \citep{cifar10}. The \textbf{MNIST} dataset comprises 70,000 grayscale images of handwritten digits (0--9), each with a resolution of $28 \times 28$ pixels, and is mostly used for classification and pattern recognition tasks due to its simplicity and accessibility. \textbf{Fashion MNIST} serves as a more challenging alternative to MNIST, containing 70,000 grayscale images of fashion items, such as shirts, shoes, and bags, also at same resolution of $28 \times 28$ pixels. This dataset provides a greater diversity in texture and structure, making it suitable for evaluating model's generalization capabilities. 

\vspace{-2pt}
As \textbf{baseline} we use a standard variational BNN \citep{blei2017variational}, 
applied to the classical Multilayer Perceptron (MLP) architecture.
The \textbf{MLP Backbone} 
is composed of 
an input layer, a single hidden layer and an output layer. The
\textbf{Input Layer} 
processes the input data with a shape corresponding to the dimensions of the dataset. For grayscale datasets (MNIST and Fashion MNIST), the input shape is $28 \times 28 \times 1$, and for CIFAR-10, the input shape is $32 \times 32 \times 3$. A
\textbf{Flattening Layer} flattens the input 
into a single-dimensional vector to be fed to the subsequent dense layers.
\textbf{DenseFlipout Layers} are implemented using TensorFlow Probability's \texttt{DenseFlipout}. 
They approximate the weight posterior distributions using a Flipout Monte Carlo estimator, which reduces the variance of gradient estimates during backpropagation. The first dense layer contains hidden units with ReLU activation, followed by a dropout layer with a rate of 0.1 to prevent overfitting. The second dense layer, which acts as the output layer, maps to the number of classes in the dataset.
The Bayesian MLP is trained using the Evidence Lower Bound (ELBO) loss function, which combines the negative log-likelihood (NLL) of the observed data with the Kullback-Leibler (KL) divergence between the approximate posterior and the prior distributions of the weights. The NLL component is computed using the softmax cross-entropy between the true labels and the predicted logits, while the KL divergence is derived from the prior and posterior distributions of the weights. The model is trained over 20 epochs (for MNIST and Fashion MNIST) and using Adam.

\vspace{-2pt}
A number of hyperparameters are fixed across all experiments: namely, 
the number of closed intervals (30) and the number of samples drawn from the posterior distributions (5,000). Additionally, the budgeting strategy is consistently applied by selecting 5\% of the weights based on the selected criteria specified per experiment.

\vspace{-2pt}
\subsection{Ablation on Budgeting}
\label{sec:ablation}
\vspace{-2pt}
We first conducted an ablation study on the MNIST dataset in which four different Budgeting criterias were tested.

\vspace{-2pt}
In \textbf{Budgeting using High Variance ($\uparrow \sigma$)} 
we sampled $5\%$ weights with 'High Variance' from the posterior distributions (parameters: $\mu, \sigma$) of the whole model and transformed them to belief posteriors using Epistemic Wrapper. The results are shown in Table \ref{combined_results},
where 
``MLP size" is the number of hidden units in the single hidden layer of the model.

\vspace{-2pt}
Since inference in our methodology is done using INNs, we compare our results with those of Hybrid INN (taken as a baseline).
The results shows that using the wrapper improves the quality of the weights initialization with respect to the Hybrid INN baseline. 
For instance, an MLP with 32 hidden units and weights randomly initialized achieved an accuracy of $10.37\%$ on the test data, while for our wrapper the test accuracy was $50.20\%$.

\vspace{-2pt}
\textbf{Budgeting using High Mean ($\uparrow \mu$)} is another strategy 
in which we sample and ``wrap" the
$5\%$ weights with 'High Mean' from the posterior distributions. 
From the results shown in Table \ref{combined_results}, it can be seen that ``High Mean" performs better for MLP size (no hidden units) $= 8$.

\textbf{In Budgeting using High Mean and High Variance ($\uparrow (\mu, \sigma$))} we rank the parameters by computing a combined score, defined as the sum of the mean and variance of their posterior distributions: $\text{combined\_score} = \mu + \sigma.$ This acts as a proxy for an upper bound of the posterior distribution, allowing us to prioritize parameters that are either highly informative (high mean) or uncertain (high variance). We then wrap these top 5\% weights using Epi-wrapper. 
The results are shown in the Table \ref{combined_results}. 
This strategy allows us to selectively wrap the most influential and uncertain parameters, ensuring that the transformation captures meaningful epistemic uncertainty. However, this approach also imposes a strict constraint on the selection process, as only weights satisfying both conditions are chosen, which may limit flexibility in certain scenarios.

\begin{table*}[!ht]
\caption{Classification accuracies on MNIST and Fashion-MNIST datasets for posterior weights transformations using `High Mean' Budgeting (5\% posterior distributions selection) before and after fine-tuning. Results are from 15 runs. MLP (Hidden Units = 8).}
\label{combined_two_datasets_results}
\vskip 0.15in
\begin{center}
\begin{small}
\begin{sc}
\resizebox{0.9\linewidth}{!}{
\begin{tabular}{cccccc}
\toprule
\multirow{2}{*}{Datasets} & \multicolumn{2}{c}{Before Fine-Tuning} & \multicolumn{3}{c}{After Fine-Tuning} \\
\cmidrule(lr){2-3} \cmidrule(lr){4-6}
 & INN & Epi-Wrapper & BNN & INN & Epi-Wrapper  \\
\midrule
MNIST   & 9.33 $\pm$ 0.54 & \textbf{51.33 $\pm$ 1.21} & 72.44 $\pm$ 0.24 & \textbf{91.12 $\pm$ 0.08} & 91.02 $\pm$ 0.05 \\
Fashion-MNIST & 8.57 $\pm$ 1.03 & \textbf{26.93 $\pm$ 1.44} & 58.91 $\pm$ 0.24 & 82.41 $\pm$ 0.19 & \textbf{82.45 $\pm$ 0.10} \\
\bottomrule
\end{tabular}}
\end{sc}
\end{small}
\end{center}
\end{table*}

\textbf{Budgeting using Random Selection (RS) ($\mu, \sigma$)} is done by randomly selecting $5\%$ weights from the baseline BNN and extract belief posteriors using the wrapper. Table \ref{combined_results} shows that the results are worse than with other strategies.
This is due to the fact that random sampling, while giving us an unbiased selection of posterior weights, may miss those posterior distributions with high uncertainty that can be improved using our wrapping approach.



\vspace{-2pt}
\subsection{Fine-Tuning}
\vspace{-2pt}

We performed the Fine-tuning of the models, the baseline (INN) and ours (Epi-Wrapper), on the training data. The results are shown in Table \ref{combined_results}. In comparison to the  INN and BNN baselines, our model performs well as the weights wrapping mechanism acts as an initialization strategy in fine-tuning.

\vspace{-2pt}
\subsection{iD and OoD Experimental Evaluation}
\vspace{-2pt}

\begin{table}[t]
\caption{Performance comparison regarding uncertainty estimation on MNIST vs Fashion-MNIST OoD samples.}
\label{Tab:OOD}
\centering
\begin{tabular}{@{}c c ccc@{}}
\toprule
                &                                                                                & \multicolumn{3}{c}{OoD Samples}                                                                                            \\ \cmidrule(l){3-5} 
                & \multirow{-2}{*}{\begin{tabular}[c]{@{}c@{}} iD Test\\ Accuracy (\%)\end{tabular}} & \multicolumn{1}{c|}{AUROC}                                             & \multicolumn{1}{c|}{AUPRC}                                             & EU                     \\ \midrule
INN             &     91.07 $\pm$ 0.08                                                              & \multicolumn{1}{c|}{0.5329}                                  & \multicolumn{1}{c|}{0.8951}                                              & 0.0023                      \\ \midrule

Epi-Wrapper & \textbf{91.12 $\pm$ 0.06 }                                                          & \multicolumn{1}{c|}{\textbf{0.6673}} & \multicolumn{1}{c|}{\textbf{0.9120}}          & \textbf{0.0059} \\ \bottomrule
\end{tabular} 
\end{table}

\begin{figure}[tbp]
    \centering
    \includegraphics[width=0.55\linewidth]{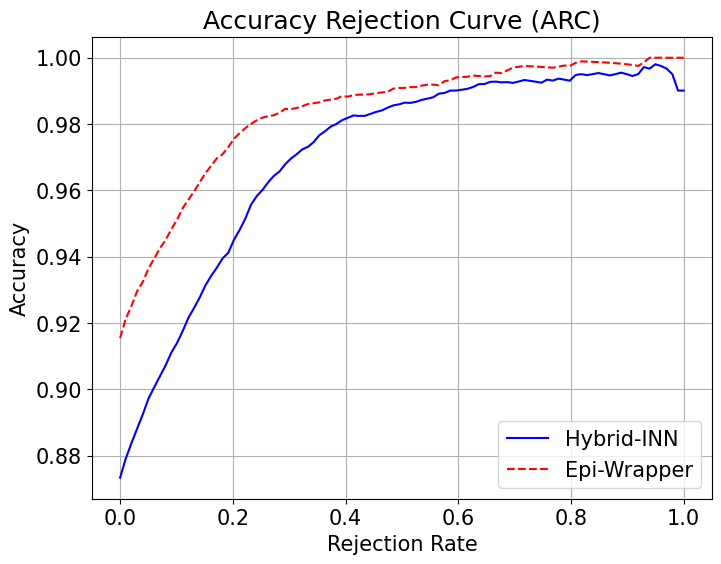}
    \vspace{-5pt}
    \caption{Accuracy Rejection Curve (ARC) for the MNIST dataset, comparing INN and Epi-Wrapper.}
    \label{fig:arc}
        \vspace{-15pt}
\end{figure}

\textbf{Analysis of Accuracy Rejection Curve (ARC) on in-Domain Data}
Figure~\ref{fig:arc} presents the Accuracy Rejection Curve (ARC) for the in-domain (iD) dataset, comparing the performance of two models, INN and Epi-Wrapper, on MNIST. In the in-domain setting, both the training and testing are conducted on the MNIST dataset, with 
no distributional shift. 
It can be seen in the plot that, as the rejection rate increases, the accuracy steadily improves for both models, demonstrating that removing high-uncertainty predictions enhances the overall correctness of classification. However, the Epi-Wrapper model consistently outperforms the INN model across the entire range of rejection rates. This suggests that Epi-Wrapper is better at identifying and filtering uncertain predictions, leading to an increase in accuracy. Since this evaluation is conducted in an in-domain setting (iD), the observed performance improvements primarily reflect how well the models handle uncertainty within a familiar data distribution. For OOD samples, the results are presented in Table \ref{Tab:OOD}. Clearly our Epi-wrapper performs well in comparison to the baseline (INN). 

\vspace{-2pt}
\textbf{Uncertainty Evaluation:}
In this work, we employ Monte Carlo (MC) Dropout to estimate predictive uncertainty by performing multiple stochastic forward passes through the model at inference time. Unlike standard deterministic models, where a single forward pass generates fixed predictions, MC Dropout enables the model to capture both aleatoric and epistemic uncertainty. The results are presented in Table~\ref{Tab:OOD}.  

\vspace{-2pt}
Table~\ref{Tab:OOD} presents a comparative analysis of uncertainty estimation performance between INN and Epi-Wrapper on out-of-distribution (OoD) samples from the Fashion-MNIST dataset, while both models were trained and evaluated in-domain (iD) on MNIST. The table provides key evaluation metrics, including iD test accuracy, AUROC (Area Under the Receiver Operating Characteristic Curve), AUPRC (Area Under the Precision-Recall Curve), and Epistemic Uncertainty. 
For iD Test Accuracy:  Both models achieve comparable in-domain classification accuracy on MNIST, with INN attaining an accuracy of 91.07\% and Epi-Wrapper slightly outperforming it with 91.12\%. 
AUROC on OoD Samples: This is a key metric for OoD detection, measuring the model's ability to distinguish between iD and OoD samples. 
A higher AUROC indicates better separation. Epi-Wrapper achieves a higher AUROC of 0.6673, outperforming INN (0.5329). This suggests that Epi-Wrapper is more effective at recognizing and distinguishing OoD samples from in-domain data.
AUPRC on OoD Samples: AUPRC provides insight into how well the model prioritizes high-confidence predictions for OoD detection. Epi-Wrapper achieves an AUPRC of 0.9120, surpassing INN (0.8951). The higher AUPRC indicates that Epi-Wrapper generates better-calibrated uncertainty estimates, ensuring that truly OoD samples receive higher uncertainty scores.

\vspace{-2pt}
EU represents the model's epistemic uncertainty measure, where higher values suggest better sensitivity to OoD data. Epi-Wrapper achieves an EU of 0.0059, which is more than twice the EU of INN (0.0023). This implies that Epi-Wrapper assigns greater uncertainty to OoD samples, making it more reliable in real-world scenarios where detecting unfamiliar inputs is crucial.

\subsection{Comparative Analysis}

Table \ref{combined_two_datasets_results}
offers a 
comparative analysis of classification accuracies on the MNIST and Fashion-MNIST datasets. Epi-Wrapper exhibits distinct performance advantages over its counterparts, BNN and INN. Before fine-tuning, Epi-Wrapper significantly outperforms INN on both datasets, achieving high accuracy of 51.33\% on MNIST and 26.93\% on Fashion-MNIST. 
This indicates our Wrapper’s superior ability to perform better on challenging data. After the fine-tuning process, while the BNN and INN models show notable improvements demonstrating their adaptability through training, the Epi-Wrapper maintains competitive, high-performance levels. 
Its superiority on the Bayesian baseline, both before and after fine tuning, confirms our hypothesis that employing higher-order, random-set representations in the parameter space is advantageous.

\vspace{-5pt}
\section{CONCLUSIONS}
\label{conclusion}\label{sec:conc}

This paper presented a novel methodology, Epistemic Wrapper, which, for the first time, extends higher-order uncertainty representation to the parameter space of neural networks.
Utilizing Bayesian neural networks 
as a baseline, our approach transforms their outputs into belief-function posteriors. This method effectively captures epistemic uncertainty, thus offering a robust, efficient, and generic approach to uncertainty quantification. 
Our experimental analysis on a BNN baseline with MLP architecture across two datasets, MNIST and Fahion-MNIST, validated the effectiveness of Epistemic Wrappers. The results demonstrated that Epi-Wrapper does generalise well. Our future work is to extend and validate the approach on larger scale networks, and develop a framework to use the wrapped weights to generate a predictive random set in the target space.

\bibliographystyle{plain}
\bibliography{references}

\newpage
\appendix
\onecolumn
\section{Appendix}

\subsection{Interval Neural Networks (INNs)}
\label{Subsec: ExistingINNs}

Traditional \textit{interval neural networks} use deterministic interval-based inputs, outputs, and parameters (weights and biases) for each node. The forward propagation in the $l^{\text{th}}$ layer of INNs is expressed as:
\begin{equation}
\textstyle
\begin{aligned}
[\underline{\boldsymbol{a}}, \overline{\boldsymbol{a}}]^{l}&\!=\!\sigma^{l}([\underline{\boldsymbol{\omega}}, \overline{\boldsymbol{\omega}}]^{l} \odot [ \underline{\boldsymbol{a}}, \overline{\boldsymbol{a}}] ^{l-1} \oplus[\underline{\boldsymbol{b}}, \overline{\boldsymbol{b}}]^{l})\\
&\!=\! [\sigma^{l}(\underline{\boldsymbol{o}} + \underline{\boldsymbol{b}}), \sigma^{l}(\overline{\boldsymbol{o}} + \overline{\boldsymbol{b}})]\ \text{with} \\
& \ \ \ [\underline{\boldsymbol{o}}, \overline{\boldsymbol{o}}]^{l} \!=\! [\underline{\boldsymbol{\omega}}, \overline{\boldsymbol{\omega}}]^{l} \odot [\underline{\boldsymbol{a}}, \overline{\boldsymbol{a}}]^{l-1},
\end{aligned}
\label{Eq: forward_inn}
\end{equation}
where $\oplus$, $\ominus$, and $\odot$ represent interval addition, subtraction, and multiplication, respectively \citep{hickey2001interval}. The terms $[\underline{\boldsymbol{a}}, \overline{\boldsymbol{a}}]^{l}$, $[\underline{\boldsymbol{a}}, \overline{\boldsymbol{a}}]^{l-1}$, $[\underline{\boldsymbol{\omega}}, \overline{\boldsymbol{\omega}}]^{l}$, and $[\underline{\boldsymbol{b}}, \overline{\boldsymbol{b}}]^{l}$ denote the interval-formed outputs of the $l^{th}$ and $(l-1)^{th}$ layers, as well as the intervals of weights and biases of the $l^{th}$ layer, respectively. 
${\sigma^{l}(\cdot)}$ is the activation function of the $l^{th}$ layer, which must be monotonically increasing. The application of interval arithmetic \citep{hickey2001interval} in \eqref{Eq: forward_inn} grants INNs the `set constraint' property. Specifically, for any $\boldsymbol{a}^{l-1}\!\in\! [\underline{\boldsymbol{a}}, \overline{\boldsymbol{a}}]^{l-1}$, $\boldsymbol{\omega}^{l}\!\in\! [\underline{\boldsymbol{\omega}}, \overline{\boldsymbol{\omega}}]^{l}$, and $\boldsymbol{b}^{l}\!\in\! [\underline{\boldsymbol{b}}, \overline{\boldsymbol{b}}]^{l}$, the constraint in \eqref{Eq: setConstraints} consistently holds.
\begin{equation}
\textstyle\boldsymbol{a}^{l}\!=\!\sigma^{l}(\boldsymbol{\omega}^{l}\!\cdot\!\boldsymbol{a}^{l-1} \!+\! \boldsymbol{b}^{l}) \!\in\! [\underline{\boldsymbol{a}}, \overline{\boldsymbol{a}}]^{l}.
\label{Eq: setConstraints}
\end{equation}
If $[\underline{\boldsymbol{a}}, \overline{\boldsymbol{a}}]$ is non-negative, such as the output of RELU activation, the calculation of $[\underline{\boldsymbol{o}}, \overline{\boldsymbol{o}}]$ in \eqref{Eq: forward_inn} can be simplified as:
\begin{equation}
\textstyle
\begin{aligned}
\underline{\boldsymbol{o}} &\!=\! \text{min}\{\underline{\boldsymbol{\omega}}, \boldsymbol{0}\}\!\cdot\!\overline{\boldsymbol{a}} + \text{max}\{\underline{\boldsymbol{\omega}}, \boldsymbol{0}\}\!\cdot\!\underline{\boldsymbol{a}} \\  
\overline{\boldsymbol{o}} &\!=\! \text{max}\{\overline{\boldsymbol{\omega}}, \boldsymbol{0}\}\!\cdot\!\overline{\boldsymbol{a}} + \text{min}\{\overline{\boldsymbol{\omega}}, \boldsymbol{0}\}\!\cdot\!\underline{\boldsymbol{a}}
\label{Eq: sim_full_o_lower_upper}
\end{aligned}.
\end{equation}
\subsection{Datasets:}
\textbf{CIFAR-10} is a collection of 60,000 color images (split into 50,000 training and 10,000 testing samples) across 10 classes, including animals and vehicles, with each image having a resolution of $32 \times 32$ pixels.\\
\textbf{CIFAR-100} consists of 60,000 color images, each of size $32 \times 32$ pixels with three RGB channels, divided into 50,000 training images and 10,000 test images. The dataset contains 100 fine-grained classes, with each class having 600 samples, making it a more challenging extension of the CIFAR-10 dataset. Unlike CIFAR-10, which includes only 10 broad categories, CIFAR-100 introduces a hierarchical structure, grouping its 100 classes into 20 superclasses based on semantic similarity. 

\subsection{Bayesian Baselines}
For baselines, we utilize two standard variational BNNs: BNNR (Auto-Encoding Variational Bayes \citep{kingma2013auto} with the local re-parameterization trick \citep{molchanov2017variational}), and BNNF (Flipout gradient estimator with the negative evidence lower bound loss \citep{wen2018flipout}). The results before fine-tuning are presented in Table \ref{table:full_cifar10_cifar100_results}.

\begin{figure}[ht]
\centering

\begin{subfigure}{0.3\textwidth}
\centering
\includegraphics[width=\linewidth]{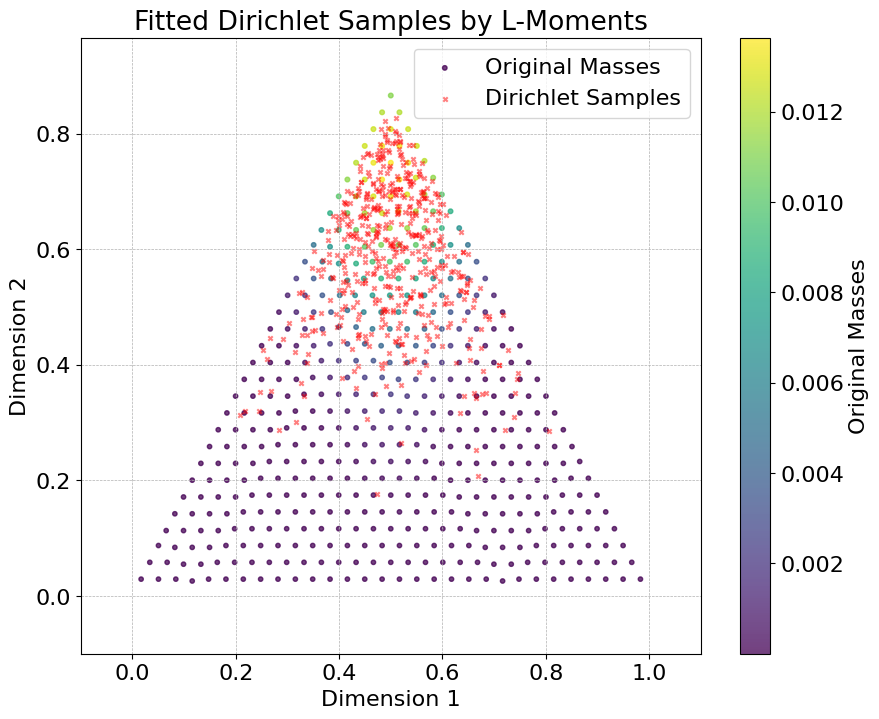} 
\caption{Number of intervals 30}
\label{fig:sub1}
\end{subfigure}
\hfill
\begin{subfigure}{0.3\textwidth}
\centering
\includegraphics[width=\linewidth]{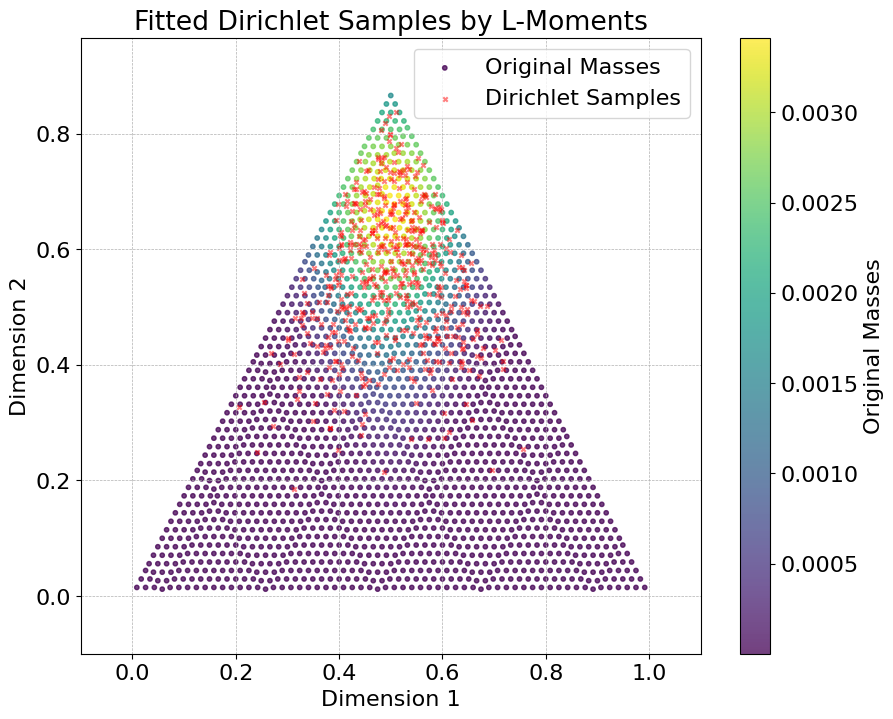} 
\caption{Number of intervals 60}
\label{fig:sub2}
\end{subfigure}%
\hfill
\begin{subfigure}{0.3\textwidth}
\centering
\includegraphics[width=\linewidth]{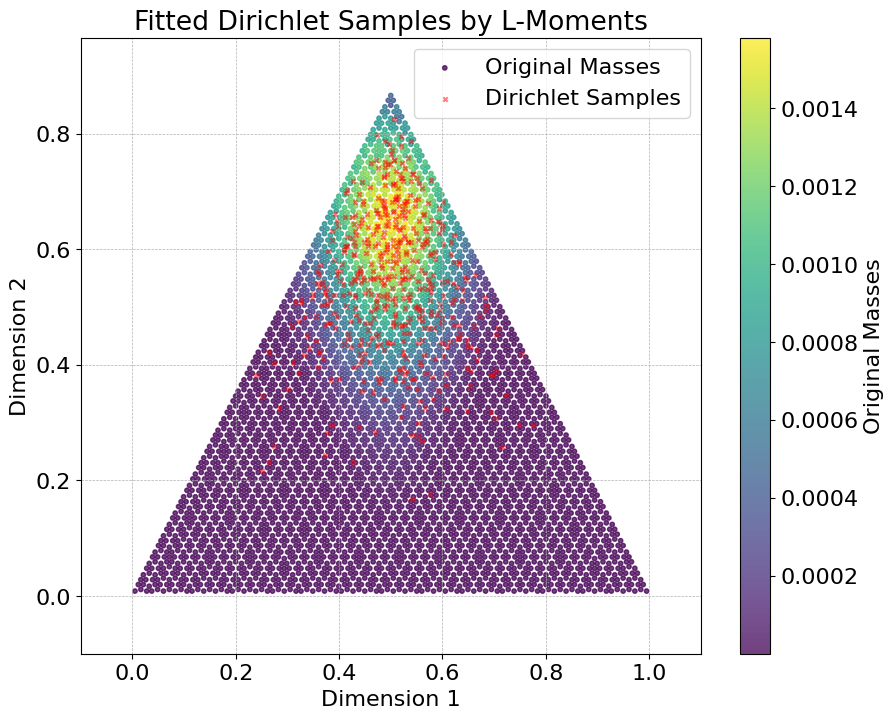} 
\caption{Number of intervals 90}
\label{fig:sub3}
\end{subfigure}

\bigskip 

\begin{subfigure}{0.3\textwidth}
\centering
\includegraphics[width=\linewidth]{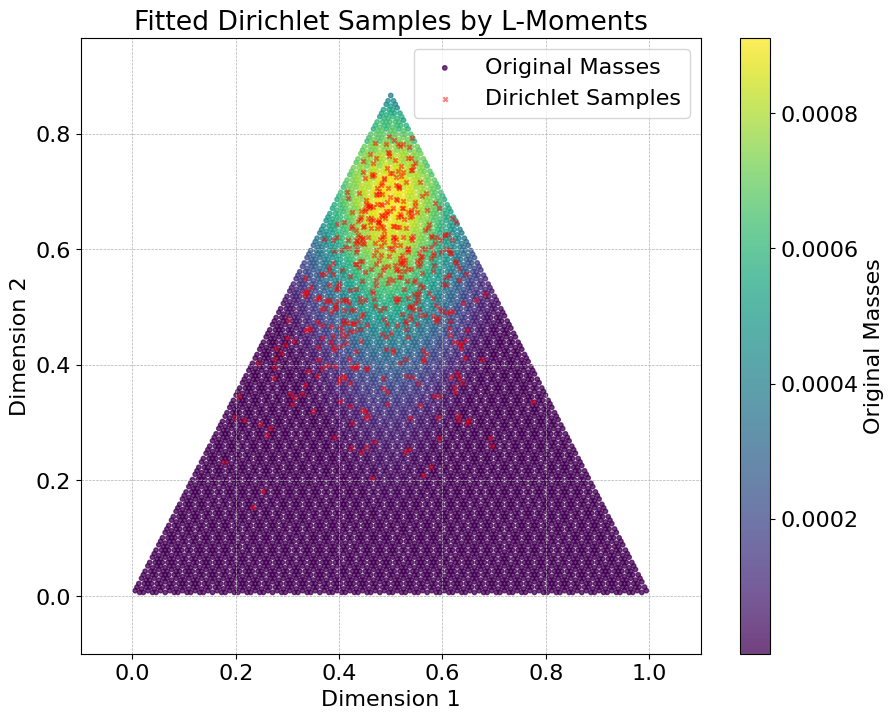} 
\caption{Number of intervals 120}
\label{fig:sub4}
\end{subfigure}%
\hfill
\begin{subfigure}{0.3\textwidth}
\centering
\includegraphics[width=\linewidth]{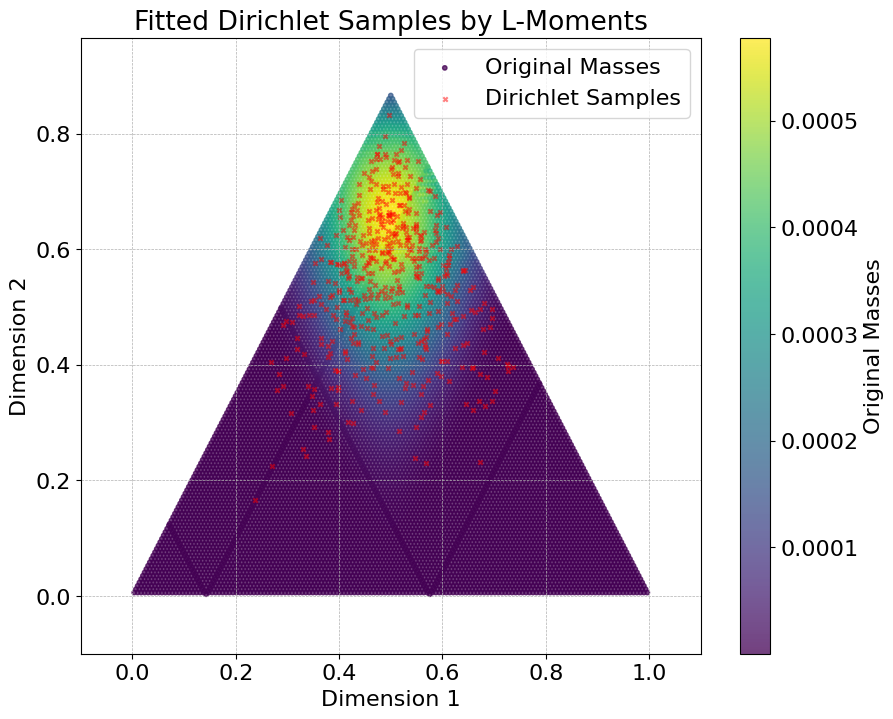} 
\caption{Number of intervals 150}
\label{fig:sub5}
\end{subfigure}%
\hfill
\begin{subfigure}{0.3\textwidth}
\centering
\includegraphics[width=\linewidth]{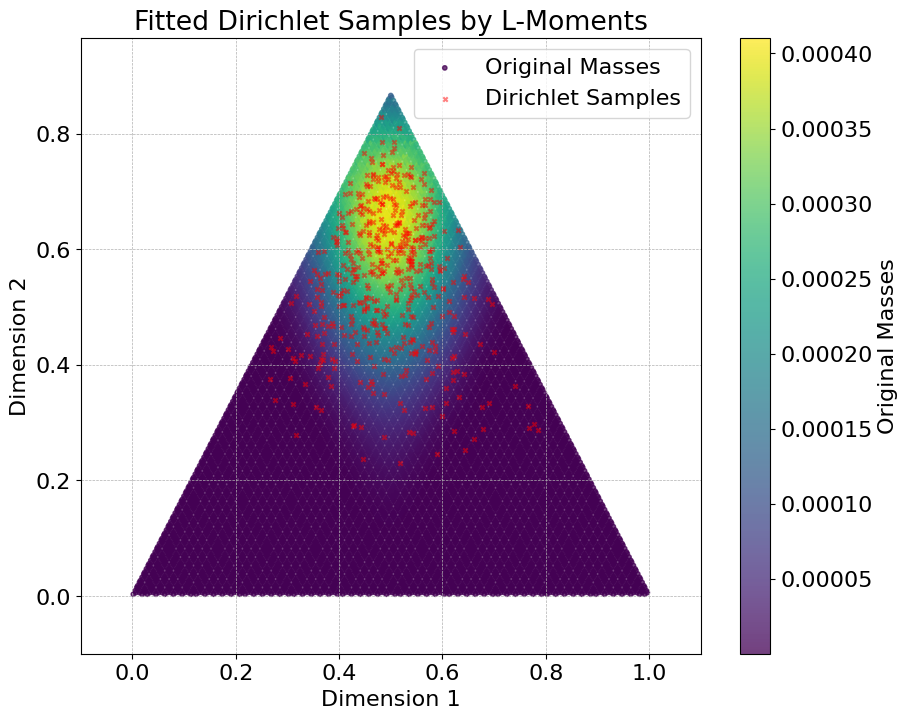} 
\caption{Number of intervals 180}
\label{fig:sub6}
\end{subfigure}

\caption{Varying Number of closed intervals}
\label{figall}
\end{figure}

\subsection{Backbones}
\textbf{LeNet-5} is adapted in this study into a Bayesian framework. Following are the details of the model's architecture. \textbf{Input Layer:} The model accepts input data with a shape corresponding to the dataset used (e.g., $28 \times 28 \times 1$ for grayscale datasets like MNIST and Fashion MNIST, and $32 \times 32 \times 3$ for RGB datasets like CIFAR-10).
\textbf{Bayesian Convolutional Layers:} The first convolutional layer employs a \texttt{Convolution2DFlipout} layer with 6 filters, a kernel size of $5 \times 5$, and ReLU activation. This is followed by an average pooling layer to reduce spatial dimensions. The second convolutional layer also uses a \texttt{Convolution2DFlipout} layer with 16 filters, a kernel size of $5 \times 5$, and ReLU activation. This layer is also followed by an average pooling layer for further downsampling.
\textbf{Flattening Layer:} After the convolutional layers, the output is flattened into a one-dimensional vector, preparing it for fully connected layers.
\textbf{Bayesian Fully Connected Layers:}
 The first fully connected Bayesian layer uses a \texttt{DenseFlipout} layer with 120 units and ReLU activation.
The second fully connected Bayesian layer uses a \texttt{DenseFlipout} layer with 84 units and ReLU activation. \textbf{Output Layer:} The output is produced by a \texttt{DenseFlipout} layer with a number of units equal to the number of classes in the dataset, applying a softmax activation function to generate class probabilities.\\

\textbf{ResNet-18} The Bayesian ResNet-18 model integrates Bayesian inference into the classical ResNet-18 architecture. This model leverages Bayesian Convolutional Neural Networks (Bayesian CNNs) with Flipout and Reparameterization layers from TensorFlow Probability, enabling weight uncertainty modeling. The architecture consists of four main residual blocks, with convolutional layers followed by batch normalization and ReLU activation. The convolutional layers employ Bayesian weight posterior distributions, where the kernel weights follow a Gaussian posterior parameterized by mean and variance. These distributions are constrained using a log-variance regularization technique, ensuring numerical stability. The weight posteriors are sampled using the Mean-Field Variational Inference approach, enabling Bayesian updates during training. The ResNet-18 backbone begins with an initial convolutional layer followed by four residual blocks, each progressively increasing the number of filters from 64 to 512. The residual connections allow gradient flow through the network, ensuring stable training. To approximate the posterior over weights, Convolution2DReparameterization and Convolution2DFlipout layers are utilized, capturing epistemic uncertainty through stochastic weight sampling. The final layers include average pooling, flattening, and a fully connected Bayesian dense layer with Flipout, producing the classification logits.\\

\textbf{VGG-16} The Bayesian VGG-16 model integrates Bayesian inference into the classical VGG-16 architecture to enable principled uncertainty estimation in deep learning. The standard convolutional layers are replaced with Bayesian Convolutional Neural Networks (Bayesian CNNs) using Convolution2DReparameterization and Convolution2DFlipout layers from TensorFlow Probability. These layers approximate posterior distributions over weights using Mean-Field Variational Inference, ensuring reliable uncertainty quantification. VGG-16 follows a deep convolutional architecture with 16 layers, consisting of multiple stacked convolutional layers with small $3 \times 3$ filters, followed by max pooling layers to progressively reduce spatial dimensions. The Bayesian adaptation maintains this structure while introducing posterior weight sampling in convolutional layers, ensuring that the feature extraction process incorporates uncertainty information. Batch normalization and ReLU activation are applied to enhance convergence stability, while Bayesian priors constrain weight posteriors, preventing overconfidence in predictions. The final classification layers include Bayesian fully connected layers with Flipout, which sample weights during inference to produce uncertainty-aware predictions.

\subsection{Hyperparameter Analysis}
The main manuscript experimental setup contains fixed hyperparameters such as number of closed intervals (30) and samples drawn from the posterior distributions (5,000). Also, the budgeting strategy is consistently applied by selecting 5\% of the weights based on the selected criteria specified per experiment.

\vspace{-2pt}
\begin{table*}
\caption{Before Fine-tuning: Performance comparison regarding Budgeting percentage on MNIST. Number of intervals fixed 30.}
\label{Table_HP}
\vskip 0.15in
\begin{center}
\begin{sc}
\resizebox{0.9\linewidth}{!}{
\begin{tabular}{@{}c cccccc@{}}
\toprule
& \multicolumn{6}{c}{Epi-Wrapper} \\
\cmidrule(l){2-7} 
\begin{tabular}[c]{@{}c@{}} INN\\ Accuracy (\%)\end{tabular} & \multicolumn{1}{c|}{5\% (Weights)} & \multicolumn{1}{c|}{10\% (Weights)} & \multicolumn{1}{c|}{20\% (Weights)} & \multicolumn{1}{c|}{30\% (Weights)} & \multicolumn{1}{c|}{40\% (Weights)} & 50\% (Weights)\\
\midrule
9.33 $\pm$ 0.54 & \multicolumn{1}{c|}{25.46 $\pm$ 1.57 } & \multicolumn{1}{c|}{45.34 $\pm$ 1.38} & \multicolumn{1}{c|}{42.25 $\pm$ 1.26} & \multicolumn{1}{c|}{32.62 $\pm$ 2.00} & \multicolumn{1}{c|}{19.07 $\pm$ 0.80} & 14.08 $\pm$ 0.94 \\
\bottomrule
\end{tabular}}
\end{sc}
\end{center}
\end{table*}

\begin{table*}
\caption{After Fine-tuning: Performance comparison regarding Budgeting percentage on MNIST.}
\label{Table_HP_afterFT}
\vskip 0.15in
\begin{center}
\begin{sc}
\resizebox{0.9\textwidth}{!}{
\begin{tabular}{@{}c cccccc@{}}
\toprule
& \multicolumn{6}{c}{Epi-Wrapper} \\
\cmidrule(l){2-7} 
\begin{tabular}[c]{@{}c@{}} INN\\ Accuracy (\%)\end{tabular} & \multicolumn{1}{c|}{5\% (Weights)} & \multicolumn{1}{c|}{10\% (Weights)} & \multicolumn{1}{c|}{20\% (Weights)} & \multicolumn{1}{c|}{30\% (Weights)} & \multicolumn{1}{c|}{40\% (Weights)} & 50\% (Weights)\\
\midrule
91.12 $\pm$ 0.08  & \multicolumn{1}{c|}{91.08 $\pm$ 0.09} & \multicolumn{1}{c|}{91.83 $\pm$ 0.04} & \multicolumn{1}{c|}{91.84 $\pm$ 0.04} & \multicolumn{1}{c|}{91.85 $\pm$ 0.13} & \multicolumn{1}{c|}{91.82 $\pm$ 0.09} & 91.50 $\pm$ 0.05 \\
\bottomrule
\end{tabular}}
\end{sc}
\end{center}
\end{table*}

\subsection{Fine-tuning on Large-scale models}
We modulate the spread of Bayesian neural network weights by applying a layer-specific scaling factor $k$ to the standard deviation when constructing interval bounds. This technique serves to regulate the initial uncertainty of model parameters. A similar concept exists in variational Bayesian inference, where a \emph{tempering parameter} $\tau$ is introduced to control the contribution of the likelihood in the posterior distribution~\cite{osawa2019practical}. Lower values of $\tau$ (or equivalently, of $k$ in our case) lead to sharper, more concentrated posteriors, enhancing training stability and convergence. This calibration mechanism ensures a well-behaved uncertainty estimate, particularly important in interval neural networks (INNs) where the width of intervals directly influences both prediction confidence and robustness.

\end{document}